\documentclass[conference]{IEEEtran}
\IEEEoverridecommandlockouts
\usepackage{cite}
\usepackage{amsmath,amssymb,amsfonts}
\usepackage{graphicx}
\usepackage{textcomp}
\usepackage{xcolor}
\usepackage{cite}
\usepackage{booktabs}
\usepackage{diagbox}
\usepackage{tablefootnote}
\usepackage{footnote}
\usepackage{multirow}
\usepackage[ruled]{algorithm2e}
\usepackage{algpseudocode}
\usepackage{amsmath}
\usepackage{graphics}
\usepackage[utf8]{inputenc}
\usepackage{epsfig}
\newtheorem{Def}{Definition}
\newtheorem{Pro}{Problem}
\newcommand{\tabincell}[2]{\begin{tabular}{@{}#1@{}}#2\end{tabular}}

\usepackage{subfigure}
\usepackage{helvet}  
\usepackage{courier}  
\usepackage{url}  
\usepackage{graphicx}  
\usepackage{makecell}
\usepackage{footnote}
\usepackage{tablefootnote}
\usepackage{enumerate}
\usepackage{algpseudocode}
\usepackage{bm}
\usepackage{marvosym}
\usepackage{graphics}
\usepackage{url}
\usepackage{color}
\usepackage{epstopdf}
\usepackage{subfigure}

\usepackage{breqn}
\usepackage{multirow}
\usepackage{booktabs}
\usepackage{subfigmat}
\usepackage{subfig}
\def\BibTeX{{\rm B\kern-.05em{\sc i\kern-.025em b}\kern-.08em
    T\kern-.1667em\lower.7ex\hbox{E}\kern-.125emX}}
\begin{document}

\title{Towards Learning in Grey Spatiotemporal Systems: A Prophet to Non-consecutive Spatiotemporal Dynamics
}
\author{\IEEEauthorblockN{Zhengyang Zhou$^\dagger$, Kuo Yang$^\dagger$, Wei Sun$^\dagger$,  Binwu Wang$^\dagger$, Min Zhou$^\lozenge$, Yunan Zong$^\dagger$, Yang~Wang$^\dagger${\Letter}}
	\IEEEauthorblockA{\\
		{$^\dagger$University of Science and Technology of China, Hefei, China
		}\\
		{$^\lozenge$Huawei Technologies, Shenzhen, China
		}\\
		Email:\{zzy0929, yangkuo, sunwei3, wbw1995, zyn0728\}@mail.ustc.edu.cn, zhoum1900@163.com, angyan@ustc.edu.cn{\Letter}
	}
}


\maketitle

\begin{abstract}
	Spatiotemporal forecasting is an imperative topic in data science due to its diverse and critical  applications in smart cities. Existing works mostly perform consecutive predictions of following steps with observations completely and continuously obtained, where nearest observations can be exploited as key knowledge for instantaneous status estimation. However, the practical issues of early activity planning and sensor failures elicit a brand-new task, i.e., non-consecutive forecasting. In this paper, we define spatiotemporal learning systems with missing observation  as Grey Spatiotemporal Systems (G2S) and propose a Factor-Decoupled learning framework for G2S (FDG2S), where the core idea is to hierarchically decouple multi-level factors and enable both flexible aggregations and disentangled uncertainty estimations. Firstly, to compensate for missing observations, a generic semantic-neighboring sequence sampling is devised, which selects representative sequences to capture both periodical regularity and instantaneous variations. Secondly, we turn the predictions of non-consecutive statuses into inferring statuses under expected combined  exogenous factors. In particular, a factor-decoupled aggregation scheme is proposed to decouple factor-induced predictive intensity and region-wise proximity by two energy functions of conditional random field. To infer region-wise proximity under flexible factor-wise combinations and enable dynamic neighborhood aggregations, we further disentangle compounded influences of exogenous factors on region-wise proximity and learn to aggregate them. Given the inherent incompleteness and critical applications of G2S, a DisEntangled Uncertainty Quantification is put forward, to identify two types of uncertainty for reliability guarantees and model interpretations. Extensive experiments demonstrate that our solution can promote the performance by at least 8.50\%  on early planning and 2.01\%-18.00\%  on the sensor failure setting.
	
\end{abstract}


\section{Introduction}
With the explosion of intelligent sensing devices in cities, spatiotemporal learning, which aims at supporting various smart city applications including intelligent transportation~\cite{yuan2018hetero, zhang2017deep}, smart grid~\cite{le2020probabilistic,wu2020adversarial} as well as weather forecasting~\cite{hewage2021deep,wu2021dynamic}, has become a pivotal technique for modern cities. Generally, traditional spatiotemporal forecasting tasks mostly assume that the information of urban systems is fully obtained, and data integrity has become an essential condition of the success of those traditional methods.

\begin{figure}[ht]	
	\centering
	\includegraphics[width=2.8in]{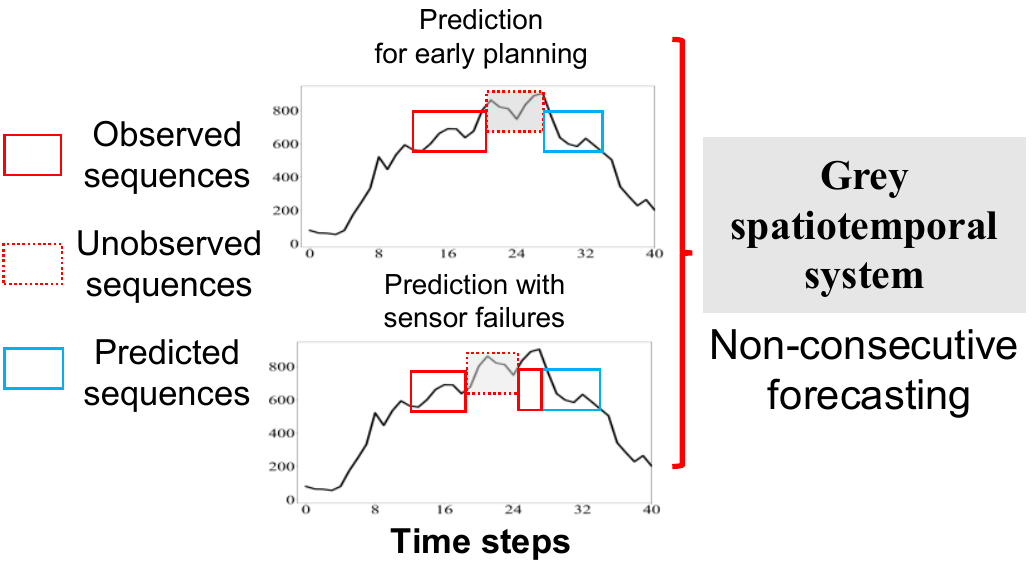}
	\caption{Illustration of grey spatiotemporal system and two typical scenarios of non-consecutive prediction }
	\label{fig:motiv}
\end{figure}

However, in real-scenarios, nature to us is not black or white, which means precise information of urban systems is totally missed or obtained. Instead of being black or white, urban systems are usually grey, i.e., with incomplete information, and this may be an inevitable obstacle of achieving various smart city applications. Specifically, there are two possible scenarios: i) The pre-scheduling of individual urban trips for vital personal events and pre-control of citywide traffics for important urban  activities take the possible urban perceptions of some day or even weeks in the future as a prerequisite; ii) With the ever-expanding deployment of urban sensors, the probability of sensing failures  increases  correspondingly, and this will definitely brings more gaps to urban sensing datasets in temporal perspective. These two scenarios both point to a new unresolved issue, spatiotemporal forecasting with fragments of unobserved sequential information, i.e., non-consecutive spatiotemporal prediction, towards a more robust urban learning system. To this end, as illustrated in Figure~\ref{fig:motiv}, we  define that urban spatiotemporal systems with fragment of observation missing as Grey Spatiotemporal System (G2S). Considering that an inherent property of grey spatiotemporal systems is the  incompleteness of data, a key issue is how to advance traditional next-step prediction to non-consecutive prediction by discovering urban regularities and disentangling potential urban structures under limited data. 

\textbf{Challenges.} Given the conflict between the requirement of traditional spatiotemporal learning methods on data integrity and the existence of continuous data gaps, the main challenges of non-consecutive spatiotemporal prediction can be summarized from the following two aspects.

\textbf{(1) Unavailability of nearest historical observations.} Due to the closeness property of spatiotemporal observations, existing prediction methods in white urban systems directly involve the observations of nearest steps as features for training, hence accurately capturing the status evolutions towards the near future~\cite{han2021dynamic,bai2019stg2seq,ye2019co,zhang2017deep}. Generally, regarding traditional prediction methods, the length of time interval determines the width of the gap between two consecutive statuses, and eventually the final prediction accuracy. Therefore,  the nearest statuses play a significant role in forecasting as they can provide key informative knowledge for status estimations on following consecutive steps. Even for those sparse spatiotemporal learning efforts where sensors are sparsely deployed in spatial domain, researchers can still take the status of spatially neighboring sensors as proxy~\cite{wang2018real} or generate real-time data with well-designed discriminators~\cite{wang2022inferring}. Unfortunately, based on above analysis, these solutions are actually an interpolation strategy that takes  great advantage of nearest observations in both temporal and spatial perspectives, and are incapable of dealing with the serial observation missing issue in our task.

\textbf{(2) Uncertainty quantification and disentanglement.} Considering scenarios of non-consecutive predictions, i.e., pre-arrangements of both crucial urban activities and vital individual travels, they are both sensitive and intolerable to significant uncertainty. But unfortunately, the data incompleteness is intrinsic for grey systems, leading to the inevitability and major concern of serious uncertainty issue in grey systems and their applications. 
On the other hand, previous literature reveals that uncertainty in a learning system can be decomposed into epistemic and aleatoric. The epistemic one accounts for the insufficiency degree of knowledge that model learned from samples, while the aleatoric one characterizes the variations induced by inherent noise and unobservable factors~\cite{kendall2017uncertainties}. Identifying these two kinds of uncertainties not only helps understand the credibility of model outputs~\cite{depeweg2018decomposition}, but also can potentially  moderate sample distributions and increase the anti-noise capacity  with respective epistemic  and aleatoric uncertainty. However, pioneering works of spatiotemporal learning with uncertainty quantification ~\cite{liu2020probabilistic,vandal2018quantifying,Wu2021quantify} are mostly modified from classification tasks and fail to explicitly disentangle two types of uncertainty. Therefore, how to provide responsible predictions with effective uncertainty measurements and explicitly disentangling two types of uncertainty is another main challenge in grey system learning and understanding.

\textbf{Present works.} In this paper, we aim to perform responsible learning in grey spatiotemporal systems by considering two non-consecutive forecasting settings, which simultaneously presents point predictions and uncertainty estimations. To this end, we propose a novel solution, Factor Decoupled Graph Learning framework for Grey Spatiotemporal System (FDG2S), integrating a Factor-Decoupled Graph Sequence Learning (FoDGSL) and a DisEntangled Uncertainty Quantification (DisEUQ) by factor decoupling strategies.
In FoDGSL, to address the unavailability of nearest historical observations, we progressively decouple multi-level factors and enable a dynamically learnable aggregation to vicariously estimate the statuses of future spatiotemporal elements. First, the contributors of target predictions are  decomposed into endogenous historical observations and exogenous context factors. For historical observations, we devise a generic semantic-neighboring sequence sampling to retrieve representative sequences of periodical regularity and instantaneous variations for pattern extraction in training models. To perform customized aggregations and enable predictions of future intervals, we resort to exogenous factors as a status proxy and turn the predictions of non-consecutive future status into inferring status with combined expected exogenous factors. Specifically, by leveraging conditional random fields, factor-decoupled aggregation is proposed to decouple factor-induced intensity and region-wise aggregations by two energy functions. Further, we disentangle influences of different exogenous factors on region-wise aggregations and learn to re-aggregate, enabling the non-consecutive predictions conditioned on arbitrary combinations of exogenous factors. To disentangle the biases regarding two types of uncertainty and boost G2S robustness, a DisEUQ is proposed by considering factors of model, sample and context. In DisEUQ, a post-explained sample density prober estimates the epistemic uncertainty from sample-specific variances. By re-utilizing and organizing exogenous factors, our weak-supervised aleatoric variation learner can infer the aleatoric uncertainty and suppress the effects of outlier samples, providing responsible predictions and grey system understanding. In summary, the following \textbf{contributions} are made in this work.

\begin{itemize}
	\item \textbf{Novel data organizations.} To remedy lacking observations, a semantic-neighboring sequence sampling with factor-aware constraints is proposed to perform personalized sampling of inherent periodicity and instantaneous variations. We  also take exogenous factors as an intermediate proxy, and reorganize main observations by the types of exogenous factors to further achieve pluggable factor-wise combinations and factor-induced variations for future target estimations.
	
	\item \textbf{FoDGSL} {hierarchically disentangles multi-level factors 
		to enable dynamic aggregations conditioned on arbitrary factor-wise combinations. We design a sampling strategy to retrieve representative sequences, a factor-decoupled aggregation to jointly disentangle complicated factor influences on aggregations and couple both endogenous and exogenous factors into a unified graph sequence learning architecture. Finally, the LSTM captures the sequential evolution for achieving longer-horizon predictions.
		
		\item \textbf{DisEUQ} is proposed to identify two sources of uncertainty and alleviate their negative influences}. A post-explained sample density prober is devised  to explore the epistemic uncertainty regarding learning experiences, while an intrinsic aleatoric variation learner is to quantify aleatoric uncertainty and suppress effects of outliers. 
\end{itemize}

\section{Preliminaries}
\subsection{Notations and problem definitions}
{\em \begin{Def}[\textbf{Spatial region division and constructed urban graph}]
		The studied areas are discretized into a set of $N$ spatial regions $\mathcal{V} = \{ {v_1},{v_2},...,{v_N}\} $ by geographical divisions or natural observation stations, and all regions can be constructed as an urban graph $G(\mathcal{V},\mathcal{E})$. The node set $\mathcal{V}$  represents fine-grained urban regions and  edge set $	{\mathbb{E}} = \{ {e_{ij}}\left| {1 \leqslant i,j \leqslant N}, \& \right.\:i \ne j{\text{\} }}$  describes potential dependencies between  two urban regions, which can be measured by  geographical proximity or other correlation metrics.
\end{Def}}

{\em \begin{Def}[\textbf{Observations of endogenous spatiotemporal elements}]
		By discretizing time domain into an interval set  $\mathcal{T} = \{ 1,2,3...,T \}$, the endogenous spatiotemporal observations are defined as the task-specific primary elements,  $\mathbb{X} = \{  \mathbf{X}_{:,1},\mathbf{X}_{:,2},...,\mathbf{X}_{:,T} \} \in \mathbb{R}{^{N \times T}}$, where ${x_{i,t}} \in \mathbf{X}\; (t \in \mathcal{T})$ denotes the specific observation at region $v_i$ during interval $t$.  We also 
		define $T_d$ as the number of  intervals  in each day.
\end{Def}}
{\em \begin{Def}[\textbf{Observations of exogenous factors}]
		Environmental context factors that are not for predictions but beneficial to task optimization are defined as exogenous factors. Given $M$ types of exogenous factors, the meta set of factor types is denoted as $\mathbb{C} =\{\mathit{Cf}_1,\mathit{Cf}_2,\cdots, \mathit{Cf}_M\}$.   We instantiate the exogenous factors in this task, $\mathit{Cf}_d$ = day of week, $\mathit{Cf}_s$ = daily  timestamps, $\mathit{Cf}_w$ =  weather  type, $\mathit{Cf}_{nw}$ = numerical  weather   values, and $\mathit{Cf}_l$ = location  embedding for $\mathit{Cf}_1$ to $\mathit{Cf}_5$. Let ${{\mathbf{C}}_t} = \{ \mathbf{c}_m(i,t) \in {\mathbb{R}^{1 \times {d_m}}}|1 \le i \le N,1 \le m \le M\} (1 \le t \le T)$ as the specific observations of exogenous factors, where $\mathbf{c}_m(i,t)$ is the concrete value of factor $\mathit{Cf}_m$ at region $i$ during $t$, and ${d_m}$ is the dimension of $\mathbf{c}_m(i,t)$.
\end{Def}}

{\em \begin{Def}[\textbf{Grey spatiotemporal system}] The combinations of  data and algorithms focusing on spatiotemporal learning tasks are defined as spatiotemporal  systems. In this work, we then define the spatiotemporal systems with fragment of observation missing  as grey spatiotemporal systems, where the span of unobserved sequences ranges from one-day to one-month. We exclude the scenarios of large-scale sensor failures and spatiotemporal intermittent as they are respectively impossible for imputation or can be simply addressed by interpolation. 	
\end{Def}}

{\em \begin{Pro}[\textbf{Non-consecutive forecasting for learning grey spatiotemporal system}]  Understanding  the grey spatiotemporal systems is equivalent to conducting responsible non-consecutive forecasting by  performing point predictions and uncertainty quantification. We have incomplete endogenous spatiotemporal observations  $\mathbb{X}=\{{\mathbf{X}_{:,1:s}} \cup {\mathbf{X}_{:,r:T}}\}$ and corresponding exogenous factors ${\mathbf{c}_{1:s}},{\mathbf{c}_{r:T}}$,
		where ${\mathbf{X}_{:,s + 1:r - 1}}$ and ${\mathbf{c}_{s + 1:r - 1}}$ are the missing sequences. In this work, we make the assumption that the missing sequences are consecutive and the interrupted intervals $(s,r)$ satisfy  $\{ (s,r)| 1 < s < r \leqslant T+1, \; T_d\leqslant r-s \leqslant 30*T_d\}$, i.e., the span of missing sequence ranges from one day to one month. Note that $r = T+1$ especially accounts for the scenarios of early planning with nearest observations totally missing.  We aim to design a learnable function $f_{ST}$ to maximally exploit both endogenous and exogenous factors, to predict future $h$-step spatiotemporal elements ${\widehat{\mathbf{Y}}_{i,t}}$, and quantify both epistemic and aleatoric  uncertainty  $({(\widehat{\mathbf{u}}_e)}_{i,t}, {(\widehat{\mathbf{u}}_a)}_{i,t})$, where $t=\{t|T < t \leqslant  T + h\}$.
\end{Pro}}

\subsection{Preliminaries of uncertainty quantification } 
Uncertainty quantification plays a critical role in understanding grey spatiotemporal system. In engineering modelling, uncertainty can be categorized into epistemic and aleatoric~\cite{der2009aleatory}. Epistemic uncertainty, raised by unseen samples and fitting capacity, can be viewed as the model parameter variation during training period, and can be reduced by incorporating more prior information, e.g., increasing the diversity and quantity of training samples. By imposing a distribution over learned parameters, quantification of epistemic uncertainty is to estimate the parameters of this potential distribution over weights. The general solution of epistemic UQ is to impose dropout during both training and testing periods, or insert  Bayesian Neural Networks, to derive multiple outputs from specific samples, and then compute the statistical variance of these predictions. Let  ${\widehat{y}_t^2}$ be $t$-th sampled predictions and we repeat the sampling for $T$ times. The estimated epistemic uncertainty ${\widehat{u}_{epis}}$ can be formulated as~\cite{kong2020sde,lakshminarayanan2017simple}, 
\begin{equation}
{\widehat{u}_{epis}} = {var} (y) = \frac{1}
{T}\sum\limits_{t = 1}^T {\widehat y_t^2}  - {(\frac{1}
	{T}\sum\limits_{t = 1}^T {{{\widehat y}_t}} )^2}
\label{eq:epist}
\end{equation}
Then the epistemic uncertainty ${\widehat{u}_{epis}}$ describes the potential parameter variation of the specific model with trained samples, where the model encapsulates the knowledge of samples.  

Aleatoric uncertainty, stemming from inherent noise and unobservable factors in datasets, is mostly heteroscedastic and cannot be eliminated. Aleatoric uncertainty characterizes the intrinsic difficulty of learning tasks. Existing methods often construct mappings from input data to aleatoric uncertainty and maintain the consistency between errors and learnable aleatoric uncertainty with  negative log likelihood  loss~\cite{kendall2017uncertainties}. We denote $D$ as the number of samples, and the loss function ${Loss_{au}}(\theta )$ for predictive aleatoric uncertainty is,
\begin{equation}
{Loss_{au}}(\theta ) = \frac{1}
{D}\sum\limits_i {\frac{1}
	{2}} \frac{{||{y_i} - {{\widehat y}_i}|{|^2}}}
{{\widehat{u}_{a,i}^2}(x)} + \frac{1}
{2}\log {\widehat{u}_{a,i}(x)}^2
\label{eq:alea}
\end{equation}
where $\widehat{u}_{a,i}^2(x)$  is a function of input data point $x$ and represents the aleatoric uncertainty. Minimizing this loss is equivalent to constraining the consistency of changes between both prediction errors and aleatoric uncertainty. Such operation can discourage the model from predicting low uncertainty for points with high residual error, alleviating the influences of outliers in the probabilistic interpretation~\cite{kendall2017uncertainties}.

In this work, we will leverage these basic theories to realize our DisEUQ for providing responsible predictions in FDG2S.

\begin{figure*}[ht]	
	\centering
	\includegraphics[width=7.2in]{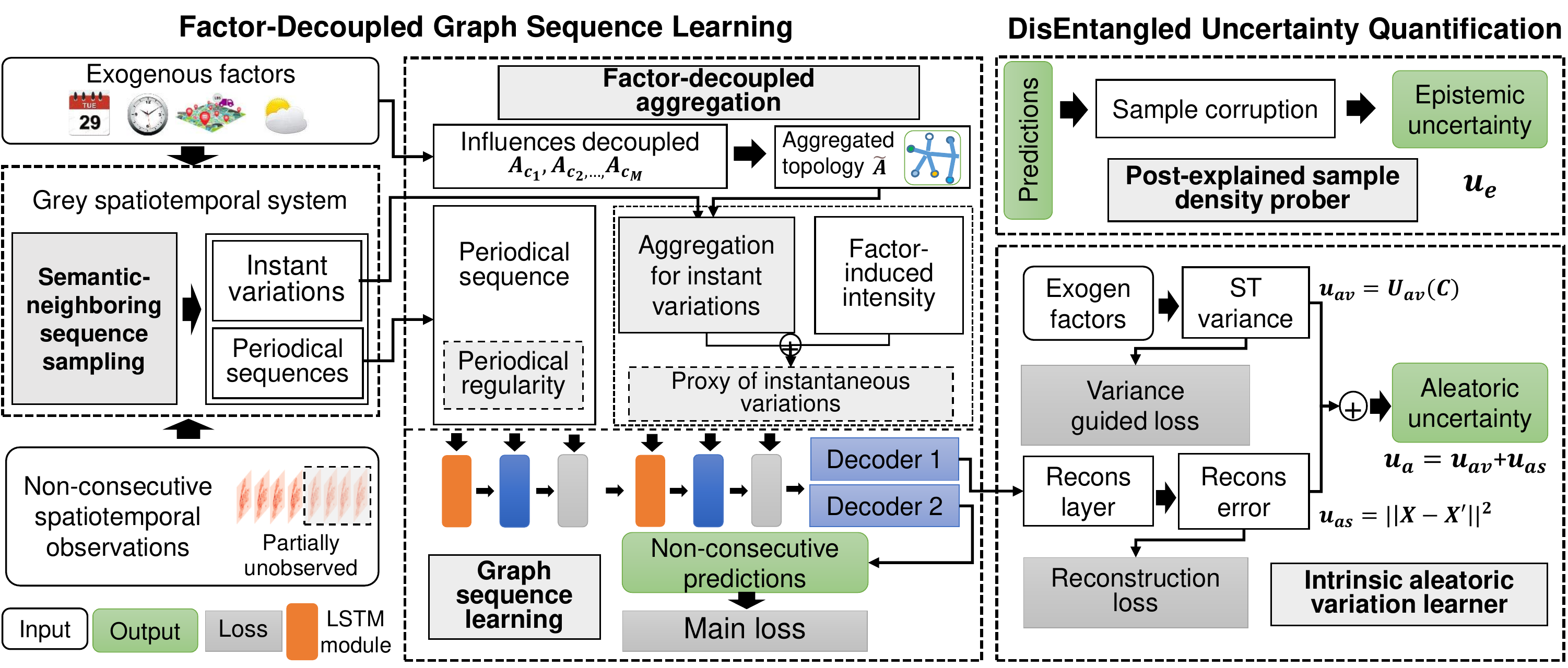}
	\caption{Framework overview of FDG2S: A factor-decoupled graph learning for grey spatiotemporal learning}
	\label{fig:FO}
\end{figure*}

\section{Methodology}
Our Factor Decoupled Graph Learning framework for Grey Spatiotemporal System (FDG2S)  is illustrated in Figure~\ref{fig:FO}, which  consists of two major components, a Factor-Decoupled Graph Sequence Learning (FoDGSL) and  a DisEntangled Uncertainty Quantification (DisEUQ). The core idea of the whole framework is to hierarchically decouple multi-level factors and enable flexible aggregations and uncertainty estimations for approaching non-consecutive predictions.


\subsection{Factor-Decoupled Graph Sequence Learning}
\subsubsection{Overview of FoDGSL} In spatiotemporal learning, the contributors of target predictions are multi-level. They can be first decomposed into two factors, endogenous historical observations and exogenous context factors. Regarding historical observations, the compositions of targeted results are distinguished by periodic regularity and instantaneous variations, while for exogenous factors, they are prone to have complex and interactive effects on targets. 
Given the intrinsic challenge of the unavailability of partial observations in grey spatiotemporal systems, the key is  to perform status inference under expected conditions by resorting to proxy. 
Fortunately, it is interesting to show that multiple exogenous factors like weather, locations, timestamps can be exploited to vicariously estimate the expected statuses of spatiotemporal elements~\cite{zhang2017deep,bai2019stg2seq,li2021spatial,Bao2019ST}. Hence, by resorting to available historical observations and informative exogenous factors, we can turn predictions of non-consecutive future statuses into  inferring status under combined expected exogenous factors. Technically, we propose an FoDGSL, to jointly consider multi-level factors and take a decoupling-and-aggregation to  enable flexible and dynamic  aggregations for approaching targets. Our FoDGSL is unified with three sub-modules, i.e., semantic-neighboring sequence sampling,  factor-decoupled aggregation to decouple and aggregate multi-level factors, and an LSTM-based graph sequence learning.

\subsubsection{Semantic-neighboring sequence sampling}
\label{sec:semantic-neigh}
To support the status estimation of unseen future, we first summarize two compositions of both periodicity regularity and instantaneous variations as the sequences that are semantic-neighboring to targets, and then select the representative periods on both aspects for pattern extractions. To simplify notations, we let $c(\mathit{Cf}_*, t)$ denote the values of exogenous factor $\mathit{Cf}_* $ at step $t$, and  we organize each sequence by $h$ steps for sequential pattern learning. First, for periodic ones, we directly search the largest $k_p$ away from $T$ that satisfies following principles,
\begin{equation}
\left\{ \begin{array}{l}
k_p \leqslant T-7*T_d \\
c(\mathit{Cf}_d, k_p) = c({\mathit{Cf}_d},T-1) \\ 
c(\mathit{Cf}_s, k_p) = c(\mathit{Cf}_s,T-1)  \\ 
{\rm IsNull}({\mathbf{X}_{ k_p - h:h}}) = {\rm False} \\ 
\end{array} \right.
\label{eq:per}
\end{equation}
We can obtain the periodical sequence $\mathbf{X}_P = \mathbf{X}_{ k_p - h:h}$, where IsNull($\cdot$) is to examine whether all elements in the function are not null. Actually, we  retrieve the intervals that simultaneously satisfy the three constraints, i.e., weekly periodical, same day of week, same index of daily intervals, to construct the temporal periodical regularity. 
We keep the retrieved daily interval $k_p$ the same  as  $T-1$, to force the model to learn sequential evolution patterns from nearest observed ones to future ones.   
Second, as the instantaneous variations usually cannot be available in G2S but could be influenced by the exogenous factors, we take exogenous factors to construct the proxy of instantaneous variations. Modifications to the former one are that 1) adding the expected weather type as an additional retrieval index to find more similar contexts in hisotrical observations, 2) removing the constraints of weekly periodicity to allow more recent available observations to be involved, 3) relaxing the same daily intervals by a tolerance value $\varepsilon $\footnote{The allowed  maximum interval shifts for tolerance $\varepsilon $ is set to 3.}. Thus, we can search the largest $k_h$ that satisfies following principles for instantaneous variations, 
\begin{equation}
\left\{ \begin{array}{l}
c(\mathit{Cf}_d, k_h) = c(\mathit{Cf}_d,T-1) \\ 
c(\mathit{Cf}_w, k_h) = c(\mathit{Cf}_w,T-1) \\ 
||c(\mathit{Cf}_s, k_h) - c(\mathit{Cf}_s,T)|| < \varepsilon  \\ 
{\rm IsNull}({\mathbf{X}_{ k_h - h:h}}) = {\rm False} \\ 
\end{array} \right.
\label{eq:hm}
\end{equation}
Then the proxy of instantaneous variations $\mathbf{X}_H$ is achieved by $\mathbf{X}_H = \mathbf{X}_{ k_h - h:h}$.
So far, the input main observations can be the ${{\rm Concat} [\mathbf{X}_P; \mathbf{X}_H]}$. Therefore, our solution is able to adaptively capture personalized semantic-neighboring sequences to support non-consecutive forecasting. When recent observations are available, this strategy degenerates to sampling the nearest consecutive sequences for instantaneous variations with Eq~(\ref{eq:hm}).

\subsubsection{Factor-decoupled Aggregation}
Since exogenous factors can significantly influence instantaneous variations thus aggregations, we are expected to realize inductive inference on non-consecutive futures by estimating the status under arbitrary combinations of exogenous factors. Technically, we propose a factor-decoupled aggregation, i.e., the decoupling-and-aggregation strategy, to progressively disentangle multiple factors and the joint influences of factors on region-wise proximity, which further learns to aggregate the proxy of instantaneous variations and imitates the evolution from available status to unseen future. We finally leverage the  expected exogenous factors to generate region-wise proximity and seamlessly couple multi-level factors  by aggregating historical observations based on such learnable adjacencies.
 
\textbf{Analysis on factor decoupling.}  In fact, the spatiotemporal elements are regularly varied with various exogenous factors for their tidal patterns and weather-induced variations, while it also reveals heterogenous region-wise proximity patterns  due to different time, weather  and holidays~\cite{bai2019stg2seq,li2021spatial,ye2019co, Bao2019ST}. In this way, the instantaneous variations influenced by exogenous factors are further decoupled into inherent intensity regularity and the factor-induced dynamic aggregations. However, the influences of combined exogenous factors on region-wise aggregations are interactive and intractable. To explicitly accommodate  the influences of dynamic factors on region-wise proximity, a factor-wise influence decoupling is motivated to disentangle and adaptively aggregate these influences for better generalization on arbitrary combinations of exogenous factors.
Therefore, the aggregations of instantaneous variation learning can be decomposed into two aspects and we resort such aggregation modelling to the theory of conditional random field (CRF). CRF is capable of capturing both mappings directly from observable factors to targets and pairwise structural correlations between reference data points~\cite{gao2019conditional}, which opportunely matches the two factors of our instantaneous variations. Accordingly, we respectively take the exogenous factors as observable variables and region-wise spatiotemporal observations as targeted variables. Then we can couple two types of CRF's energy functions, i.e., status feature function and transition feature function with graph representation learning,  to realize factor-decoupled aggregations.

\textbf{CRF-based factor-decouple modelling.} Formally, we utilize $\mathbf{c}(i)$, and ${\mathcal{N}_i}$ to denote the combined exogenous vector and the set of neighboring nodes of $v_i$, omitting the interval index for simplification,
\begin{equation}
E({y_i}|{\mathbf{c}(i)}) = {\varphi _u}({y_i},{\mathbf{c}(i)}) + \sum\limits_{j \in {\mathcal{N}_i}} {{\varphi _p}({y_i},{y_j}|{\mathbf{c}(i)})} 
\end{equation}
where ${\varphi _u}$ is the status feature function capturing the potential intensity patterns from exogenous  factors to node values, and ${\varphi _p}$ represents transition feature function modelling pairwise structural correlations  conditioned on exogenous factors. 

\textbf{Status feature function for modelling factor-induced intensity.} Given above, let $H_i$ be the $i-th$ node representation, we can construct the
contributor of combined factors to spatiotemporal observations as a  status feature function ${\varphi _u}$ of CRF, which learns factor-induced intensity and enhances exogenous factor-associated node representation,
\begin{equation}
{\varphi _u}({y_i},\mathbf{c}(i)) = ||{H_i} - B_i(\mathbf{c}(i);w_{B})||_2^2
\end{equation}
where $ B_i(\mathbf{c}(i);w_{B})$ is realized with a two-layer fully-connected network based on learnable parameters $w_{B}$.

\textbf{Transition feature function for capturing node-wise structural correlations and decoupling factor-wise influences.}
As various exogenous context factors have interactive influences on region-wise proximity thus their aggregations, we resort to these informative context factors for status estimation.
Recall that transition feature functions in CRF are capable of exploiting observable factors to quantify the node-wise proximity. We take this pairwise transition feature function for generating the region-wise proximity conditioned on exogenous factors, and we take sequence-level similarity of main observations as the region-wise proximity, where the intuition is that similar sequences of multivariate series tend to be aggregated to benefit accurate predictions. Even so, the joint influences of multiple exogenous factors on dynamic region-wise correlations are complex and intractable. To tackle this challenge, inspired by the mean-field theory that argues the status of a complex system can be considered as the summation of each related single-body system~\cite{barabasi1999mean}, we design a concise and interpretable influence decoupling mechanism. Specifically, we treat the combined exogenous factor influences on region-wise correlations as the linear combinations of single factor-induced region-wise similarity and an interactive factor. Hence, let $c_k$ denote the $k$-th category exogenous factor, our factor-decoupled pair-wise correlations can be written by modified pairwise transition feature function,  
\begin{equation}
\small
{\varphi _p}({y_i},{y_j}|{c_1},{c_2}...,{c_M}) = \sum\limits_{k = 1}^{M + 1} {{g_k}({c_1},...,{c_M})\mathit{sim}({H_{i|{c_k}}},{H_{j|{c_k}}})}
\end{equation}
where ${\varphi _p}$ describes the joint region-wise similarity conditioned on combined exogenous factor $\mathbf{c}$, $\mathit{sim}(p,q)$ measures the similarity between elements $p$ and $q$, and $g_k (\cdot)$  weighs the importance of each factor-induced similarity. Note that the $\mathit{M+1}$ category of similarity denotes the interactions of all exogenous factors that will be computed later.  To achieve single factor-induced proximity, we 
 re-organize historical observations by individual exogenous factors and resort to sequence-level similarity to surrogate node-wise similarity. Specifically, we compute node-wise sequence-level similarities conditioned on specified exogenous factor,  and average their sequence-level similarities by,
\begin{equation}
\small
sim({H_{i|{c_k}}},{H_{j|{c_k}}}) = \frac{1}
{{| \mathcal{T}_{c_k}|}}\sum\limits_{ts \in \mathcal{T} _{c_k}} {sim({H_{i,ts|{c_k}}},{H_{j,ts|{c_k}}})}
\label{eq:simsep} 
\end{equation}
where $ts \in \mathcal{T}_{c_k}$ referring to the set of timestamps satisfying that the exogenous factor category of statuses are $c_k$, and  $sim$ can be instantiated as the average of Euclidean distance and cosine similarity to preserve  both intensity and trend property. 
 Based on above, we can derive the interactions among all exogenous factors by the continued product, 
\begin{equation}
\mathit{sim}({H_{i|{c_{M + 1}}}},{H_{j|{c_{M + 1}}}}) = \prod\limits_{k = 1}^M {\mathit{sim}({H_{i|{c_k}}},{H_{j|{c_k}}})} 
\end{equation}
Essentially, our strategy can be viewed as optimizing the maximum conditional probability of region-wise adjacencies, given available exogenous factors and historical observations.

\textbf{Learning to aggregate.} 
To adapt various interaction effects induced by different factors, we devise to learn to aggregate these influences by a learnable vector $\overrightarrow g  = [{g_1},{g_2},...,{g_{M + 1}}] \in {{\mathbb R}^{M + 1}}$, enabling the single context influences to change dynamically with their varied combinations. By denoting the single context $c_k$-induced region-wise similarity ${sim({H_{i|{c_k}}},{H_{j|{c_k}}})}$ as ${{{\widetilde A}_{|{c_k}}}}$, the objective of energy function ${\varphi _p}$ is equivalent to optimizing the following region-wise conditional similarity  regarding the whole graph for target predictions. Thus, we can obtain the factor-decoupled adjacent matrix for message propagation as,
\begin{equation}
\small
{\widetilde A_{|\{ {c_1},...,{c_M}\} }} = \sum\limits_{k = 1}^M {{g_k}{{\widetilde A}_{|{c_k}}} + {g_{M+ 1}}\prod\limits_{k = 1}^M {{{\widetilde A}_{|{c_k}}}} } 
\label{eq:aggre}
\end{equation}
 Then ${g_k}$ can be computed by imposing a series of learnable parameters ${S_k}$,
\begin{equation}
\small
{g _k} = \frac{{\exp ({S_k}^T{c_k})}}
{{\sum\limits_{k = 1}^{M + 1} {\exp ({S_k}^T{c_k})} }}
\end{equation}
Here, $c_{M+1}$ is the factor-wise interaction combined by ${c_{\mathit{M + 1}}} = \mathop{\rm Concat}\limits_k \{{c_k}\} (1 \leqslant k \leqslant M)$. 

Our Factor-Decoupled Aggregation can be viewed as introducing two new additional objectives into the end-to-end graph representation learning by borrowing the idea of CRF. 
In particular, the factor-decoupled pair-wise correlation matrix  ${\widetilde A_{|\{ {c_1},...,{c_M}\} }}$ serves as the modified adjacence and the factor-induced intensity energy function  $\varphi _u (y_i, \mathbf{c}(i))$ is leveraged to enhance context factor-related representations. 
The retrieved sequence periods $X_{H}$ are  expected to feed into our graph learning for information propagation, and we formulate the $l$-th layer of our message propagation by GNN,
\begin{equation}
H_i^{(l)} = \alpha B_{_i}^{(l-1)}(\mathbf{c}(i)) + (1-\alpha) \sum\limits_{j \in {N_i}} {{{\widetilde A}_{(i,j)|\{c_1,...,c_M\}}}{X_{H}(i)}\omega _{{G_i}}^{(l-1)}} 
\end{equation}
where  $\omega _{{G_i}}^{(l)}$ are a series of layer-wise parameters for GNN aggregation. Hyperparameter $\alpha$  adjusts the importance between two contributors\footnote{$\alpha$ is initialized as 0.5 and we fine-tune them in the experiments.}.  
We stack two Factor-Decoupled GNN layers to prevent oversmooth issue, and obtain the $h$-step representation for the  proxy of aggregated instantaneous variations  ${\mathbf{H}^{{ins}}} \in \mathbb{R}^{N\times h}$.

\subsubsection{Graph sequence learning } Finally, we cascade LSTM layers with Factor-decoupled aggregation, to empower our FoDGSL to capture sequential patterns. As the periodical composition is mostly stable while the instantaneous variations tend to transit their current status to future ones, we concatenate the $h$-step  periodicity sequence $\mathbf{X}_P$  and the after-aggregated instantaneous variations  ${\mathbf{H}^{{ins}}}$ of $h$ steps into image-like sequences, and jointly feed them into a sequence learning scheme for carefully extracting temporal regularity, realizing the couplings of various level factors. 
\begin{equation}
\widehat{\mathbf{Y}}_{:,T + t} = {\rm LSTM}(({\mathbf{X}^{P}},{\mathbf{H}^{ins}}), \mathbf{W}_{lstm})
\end{equation}
where the {\rm LSTM} takes $2h-$step sequences as inputs and outputs an $h-$step predictions for non-consecutive forecasting in G2S, $\mathbf{W}_{lstm}$ are learnable parameters.
Thanks to the factor-decoupled graph adjacencies, representation vectors fed into LSTM preserve the factor-wise influences and thus we can take multiple exogenous factors for long-horizon status proxy, mitigating the issues of missing recent observations.

\subsection{Disentangled uncertainty quantification}
\subsubsection{Motivation} Due to the data incompleteness and crucial applications of grey spatiotemporal systems, uncertainty quantification is an essential guarantee for reliability.  Regarding two types of uncertainty, the epistemic one is the model uncertainty summarized by distribution of already learned samples and can be reduced by increasing both training samples and iterations. Hence, quantifying epistemic uncertainty  helps identify samples that are out-of-distribution (OOD) or reciprocal for model generalization. The aleatoric one, tend to be raised by inherent noise and unobservable factors. In particular, different exogenous factors influence human behaviors thus induce various possibility of accidential and unobservable events, we therefore partially attribute the aleatoric one to the exogenous factors. Quantifying aleatoric uncertainty can understand the inherent challenge of task itself, while regularizing such uncertainty can  alleviate the influences of outliers~\cite{kendall2017uncertainties}. 
Given different roles of uncertainty on models, we propose our DisEntangled Uncertainty Quantification (DisEUQ) by considering factors of model, sample and context. Specifically, DisEUQ designs a post-explained sample density prober for  exploring sample-specific epistemic uncertainty regarding experiences learned from training samples. It further re-utilizes the exogenous factors and organizes the variations of observations as the indicators of unobservable aleatoric uncertainty. Thus, an intrinsic aleatoric variation learner with self-supervised noise detection and factor-induced variation learner is proposed to quantify the aleatoric uncertainty  and suppress effects  of outliers.

\subsubsection{Post-explained sample density prober}
Epistemic uncertainty can be considered as the joint property of the model and sample distributions, and we can quantify it without modifying the already learned models. 
In this work, we argue that the epistemic uncertainty of a specific sample can be interpreted as identifying whether the knowledge of similar samples has been sufficiently learned by the model, which is equivalent to quantifying the density of samples that are similar with tested one in the training set. In this way, we name such density as local sample density, and design a sample density prober to understand it. Given the learned model ${f_{ST}}$ and specific tested sample ${X}_0$, we realize the sample density prober with a corruption-computation strategy. 
In particular, we first impose $J$ times of corruptions on the tested sample ${X}_0$  to generate the corrupted sample set with  different small ${\varepsilon _j} \ll {X_j} $~\footnote{As epistemic uncertainty is a relative value measuring model experiences during training process, the absolute value of $\varepsilon _j$ is orthogonal to our results.}, 
\begin{equation}
\widetilde{{\mathbb{X}}} = \{ \widetilde{{X_j}}|\widetilde{{X_j}} = {X_0} + {\varepsilon _j},j = 1,2,...J\} 
\end{equation}
We then derive the corrupted prediction set through ${f_{ST}}$, i.e., $
{\{ \widetilde{Y_j}\}} = {f_{ST}}(\{\widetilde{{X_j}}\})$.
Intuitively, for a tested sample, the sparser of its similar training samples will contribute to less learning experiences of the model, thus yielding larger epistemic uncertainty. Therefore, samples with larger epistemic uncertainty will reveal greater variations even encountering small corruptions. Then, we can compute the variance of the corrupted prediction results as the epistemic uncertainty ${u_{e}}$, 
\begin{equation} 
{u_{e}} = \mathbb{E}(\{\widetilde {Y_j^2} \}) - {\mathbb{E}^2}(\{{\widetilde {Y_j}}\})
\end{equation}
where $ \mathbb{E}(\cdot)$ is the statistical expectation.

Our sample density prober imitates multi-outputs of specific samples for variation computation that is equivalent to approximating the posterior distributions over model parameters~\cite{kendall2017uncertainties}.  More importantly, our strategy is free from increasing additional gradient propagations and training loads, leading to a more efficient uncertainty estimation where the computation load is only $\mathcal{O}(N)$.

\subsubsection{Aleatoric variation learner}
Aleatoric uncertainty characterizes the inherent learning challenge of the task itself where it can be explained by two aspects, i.e., potential noise in input observations, and interventions of unobservable factors. However, these two factors are intractable to quantify due to the lacking  of explicit supervision. To tackle this challenge, we resort to the exogenous factors for reflecting the unobservable uncertainty, and propose an aleatoric variation learner, to estimate the potential noise and exogenous factor-induced variations.

\textbf{Self-supervised noise detection.} Considering the noise component, the observation $X_0$ can be decomposed as ${X_0} = \widehat{X}_0 + {n_0}$, where $\widehat{X}_0$ is the ground-truth value, and we are aimed at quantifying noise ${n_0}$. Fortunately, neural networks enjoy the capacity of approximating non-linear functions, thus they are capable of capturing the regularity of dataset itself. Inspired by the autoencoder-based solutions to video anomaly detection~\cite{wu2021weakly}, we design a decoder  connected with the factor-decoupled aggregations to reconstruct the observations of instantaneous variations $\mathbf{X}_{H}$. Once the general regularity is learned by the network, the reconstruction error can be indicated as the potential noise in input sequences. We formulate the reconstruction loss as partial aleatoric uncertainty by,
\begin{equation}
{\widehat {u_{as}}} = L_{rec} = ||\mathbf{X}_{H} - \bf{\rm{Recon}} (\mathbf{X}_{H};\mathbf{W}_{rec})||^{2}_{2}
\end{equation}
where ${\rm Recon}$ is a learnable function parameterized by $\rm \mathbf{W}_{rec}$, and it can be instantiated as an MLP or LSTM~\footnote{Here we employ an LSTM layer in our work.}. 

\textbf{Weakly supervised exogenous variation learner.}
As discussed, the uncertainty induced by unobservable factors can be reflected by different exogenous factors and factor-wise interactions. We thereby propose a factor-induced variation indicator, which summarizes variations among similar contexts and serves as a weak-supervised pseudo label to reflect variations induced by unobservable factors.  Concretely, by instantiating ${d_u},{s_k},{w_j}$ as three exogenous factors regarding $u$-th day of week, $k$-th day timestamp and $j$-th weather type, the retrieved exogenous factor combinations for variation computation  is $\mathit{CF}({d_u},{s_k},{w_j}) = \{ \mathit{Cf}_1 = {d_u},\mathit{Cf}_2 = {s_k},\mathit{Cf}_3 = {w_j}\}$. The context-similar observation set is constructed by finding a series of intervals $ \mathbb{Q}= \{   {t_q}\}$ where their statuses share similar exogenous factor combinations as $({d_u},{s_k},{w_j})$. For one specific location $v_i$ and step $t$, we limit the maximum cardinality of $ \mathbb{Q}$ to ${\pi _Q}$ to indicate variations and avoid  high computing costs, then the constructed set will be,
\begin{equation}
\begin{split}
D({v_i},t)_{|{d_u},{s_k},{w_j}}  =   \{ {X_{i,t_q}}|& {c_1}(i,t_q)  =  {d_u},\;  {c_2}(i,t_q) = {s_k}, \\
& {c_3}(i,t_q) = {w_j}, t_q \in \mathbb{Q}\}
\label{equ:Dvi}
\end{split}
\end{equation}
Thus, we can derive the weakly supervised variation specified by the factor combinations, through  computing the standard variance (${\rm std}$) of set $D({v_i,t})$,
\begin{equation}
{({u_{av}})_{i,t|{d_u},{s_k},{w_j}}} = {\rm std}(D({v_i,t}))
\end{equation}
We then formulate the estimated exogenous variation as the function of combined factors, parameterized by  $\omega _{av}$,
\begin{equation}
{({\widehat u_{av}})_{i,t}}  = {\rm ReLU}({\omega _{av}}*\mathop {\rm Concat}\limits_{m \in \{ 1,2,...,M\} } \{ {c_m}(i,t)\} )
\end{equation}
The guidance for exogenous variance learning can be realized by minimizing the difference between $\widehat u_{av}$ and  $u_{av}$ as, 
\begin{equation}
{L_{av}} = {({u_{av}} - {\widehat u_{av}})^2}
\end{equation}
Hence, the above strategies enable our framework to learn the mappings from exogenous factors to potential factor-specific variations.
We can finally obtain the learned aleatoric uncertainty from two perspectives by ${\widehat u_a} = {\widehat u_{as}} + {\widehat u_{av}}$.
To strive a tradeoff between the weak supervision indicator and the factual property of uncertainty, followed by~\cite{depeweg2018decomposition}, we further insert an uncertainty-error consistency constraint ${L_{cons}} = \frac{{{{({y_i} - \widehat{y}_i)}^2}}}
{{{{(({\widehat u_{as}})_{i,t} + ({\widehat u_{av}})_{i,t})}^2}}}$ as the third term.  For one node $v_i$, the total aleatoric uncertainty learning is optimized by, 
\begin{equation}
{L_{Ale}}({X_i},{y_i},\widehat{y}_i) = {\gamma _1}{L_{rec}}({v_i}) + {\gamma _2}{L_{av}}({v_i}) + {\gamma _3}{L_{cons}}
\end{equation}
where $\gamma_k (k=1,2,3)$ are parameters balancing three losses.


\subsection{Optimization}
The main objectives of our spatiotemporal representation learning are two-fold, main loss for spatiotemporal forecasting,  and aleatoric uncertainty loss for  potential variation learning. We can easily integrate them as follows,
\begin{equation}
Loss({X_i},{y_i},\widehat{y}_i;t) = {\rm \mathit{\rm MAPE}}({y_i},\widehat{y}_i;t) + {L_{Ale}}({X_i},{y_i},\widehat{y}_i;t)
\label{eq:loss_s}
\end{equation}
where $t \in [T + 1, T + h]$. For uncertainty quantification, we have ${({\widehat u_{e}})_{i}}$ for epistemic one and ${({\widehat u_{a}})_{i}}$ for aleatoric one.

During training process, to address the intractable hyperparameter issues of task-wise weights, we adopt an adaptive weighting strategy by computing the ratio of  main loss (MAPE) to respective auxiliary loss, e.g., we initialize ${\gamma _1} = {\rm MAPE}({y_i},\widehat{{y_i}};t)/{L_{rec}}({v_i})$ in the first batch, and then  dynamically tune the weights according to above ratio in each epoch, reducing the workloads of parameter adjustments.
We finally employ the Adam optimizer~\cite{kingma2014adam} for optimization.

\subsection{Model  discussions and interpretations}
\textbf{Benefits and insights of FoDGSL.} 
The core idea of FoDGSL is decoupling and learning to  re-aggregate, and the insights are two-fold. First, the  disentangled proximity  explains the compositions of factor-wise effects in a complex system,  which enables factor-level flexible re-arrangements and combinations of various conditions on arbitrary future predictions.
Second, the ingenious combinations of GNN and CRF empower our framework to leverage informative exogenous factors to compensate for missing observations. We provide insights into coupling traditional machine learning or energy functions in other fields (e.g., physics) with DNNs to cooperatively  tackle the challenges that cannot be well-addressed by neural networks only, such as data limitations.

\textbf{Benefits and insights of DisEUQ.} Regarding the epistemic one, the benefits are two-fold, 1) high inference efficiency without additional training process, and 2) interpretation of sample-specific confidence regarding models, which enables  to moderate sample distributions and improves generalization by excluding or augmenting samples with high epistemic uncertainty.  The aleatoric uncertainty quantification explains the task-oriented learning difficulty and boosts the model robustness in two aspects. i) spatiotemporal autoencoder naturally  possesses the property of anti-noise. ii) the error-uncertainty consistency allows the network to adapt the residual's weighting, and thus suppress high uncertainty points, enabling to ignore potential outliers. 

\textbf{Efficiency issue.} In our work, we turn the CRF estimation into computing the separated similarities organized by exogenous factors in Eq (\ref{eq:simsep}) and re-aggregations. The solution contributes to the computation costs of $\mathcal{O}(M\cdot N^2)$ and $\mathcal{O}(M)$ where $M \ll N$.  Fortunately, the former computation can be done only once during non-training phrase, and leads to very limited external workloads to training process.  For UQ, our epistemic uncertainty is computed without any propagation process and achieves $\mathcal{O}(N)$ computation, while we limit the forward search times for constructing observation set of variances to ${\pi _Q} = 4$, which strives for a tradeoff between representativeness and efficiency. The whole complexity is generally identical to other GNNs and additional complexities are tolerable in our non-consecutive forecasting.

\textbf{Limitations.} First, even though  our model still requires at least two complete periodicity-based observations, leading to data acquisition limitations. Second, we design our model based on two scenarios,  however, for the arbitrary sequence missing patterns, it could be more efficient to build an adaptive model to accommodate different observation missing patterns without re-training. These two limitations can be potentially addressed by the emerging techniques of OOD learning and domain adaptations,  which are left as our future works.

\section{Experiments}
\subsection{Dataset descriptions}
We collect three spatiotemporal datasets from different cities, including Suzhou Industry Park  Surveillance (SIP), taxi trip records of New York City (NYC)~\footnote{https://www1.nyc.gov/site/tlc/about/tlc-trip-record-data.page}, and highway loop detectors of Los Angeles (Metr-LA)~\footnote{https://github.com/liyaguang/DCRNN}.
The detailed descriptions and statistics of our datasets are figured in Table~\ref{tab:dataset}. 
Regarding  exogenous context factors, the weather information is collected from API: https://api.weather.com, and  other factors including day of week, daily timestamps as well as holiday indicators  are  manually created from calendar.

\begin{table}[]
	\centering
	\caption{Dataset  statistics (m: million, k: thousand)}
	\small
	\begin{tabular}{ccccc}
		\toprule
		\hline
		Dataset & \tabincell{c}{Category \\ of datasets} & {\tabincell{c}{ \# of \\ records}} & \tabincell{c}{Time  Span} & {\tabincell{c}{\# of \\ regions }} \\ \hline
		\multirow{2}{*}{SIP} & Surveillance & 2.7 m & \multirow{2}{*}{\tabincell{c}{01/01/2017-\\03/31/2017}} & \multirow{2}{*}{108} \\ \cline{2-3}
		& Weather & 4.3k &  &  \\ \hline
		\multirow{2}{*}{NYC} & Taxi  trips & 7.5 m & \multirow{2}{*}{\tabincell{c}{01/01/2017-\\05/31/2017 }} & \multirow{2}{*}{354} \\ \cline{2-3}
		& Weather & 7.4k &  &  \\ \hline
		\multirow{2}{*}{\tabincell{c}{Metr-LA}} & {\tabincell{c}{Loop \\ detectors}}  & 4.9 m & \multirow{2}{*}{\tabincell{c}{03/01/2012- \\ 06/30/2012}} & \multirow{2}{*}{207} \\ \cline{2-3}
		& Weather & 5.7k &  &  \\ \hline
		\bottomrule
	\end{tabular}	
	\label{tab:dataset}
\end{table}

\subsection{Implementation details}
\textbf{Basic settings.} The dataset is divided by 60\%, 10\% and 30\% for training, validation and testing. We use a fixed 30-min time interval and organize samples to period levels where each period consists of $h$=6 intervals. 
During training, following Sec~\ref{sec:semantic-neigh}, we sample two semantic-neighboring sequences as input features. Regarding exogenous context factors, we sample the expected combined exogenous factors to feed into our framework. Note that we only compare the errors of samples that have the same weather context with targets.

\textbf{Exogenous factors.} We encode the exogenous contexts into fixed-length vectors and randomly initialize the location embeddings where the embeddings can be trainable during our end-to-end learning process. For fairness, we feed the same available exogenous factors into those baselines if they are with placeholders of context factors.  Otherwise, we impose an additional fully-connected layer on exogenous factors and perform element-wise additions to mainstream outputs to incorporate  exogeneous context factors. 

\begin{figure}[!]
	\centering
	\includegraphics[scale=0.9]{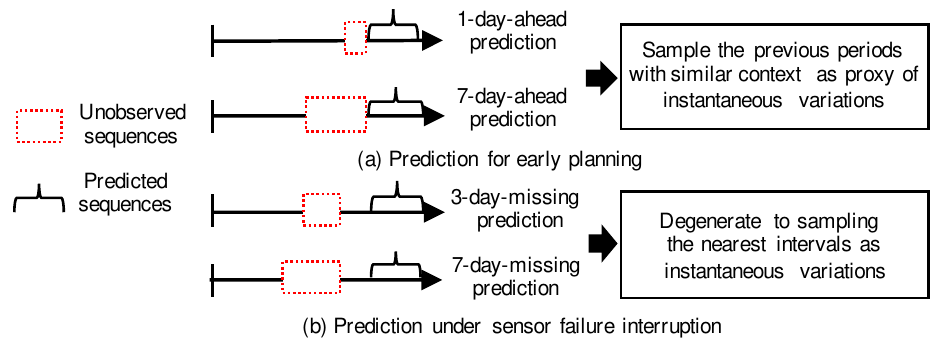}
	\caption{Illustration of non-consecutive prediction settings for grey spatiotemporal systems.}
	\label{fig:exp1}
\end{figure}

\textbf{Non-consecutive predictions.} We consider the  grey spatiotemporal systems as non-consecutive prediction tasks under the following two settings, which are illustrated in
Figure~\ref{fig:exp1}. (a) Prediction for early planning, with one-day ahead and 1-week ahead.   (b) Prediction under sensor failure interruption. To imitate scenario (b),  we fix the nearest 3-day observations available, and respectively assume that the middle 3-day and 7-day observations are missing before the nearest available 3-day observations. They are both implemented by adjusting the spans and positions of unavailable sequential observations.

\textbf{Uncertainty quantification.}
First, the sample density prober produces multiple outputs for specific samples and further derives the variance of prediction results as the epistemic uncertainty. Second, the aleatoric uncertainty is quantified and integrated by identifying potential noise and factor-induced variations, and constrained with a consistency loss function.

\subsection{Baselines}
\subsubsection{Spatiotemporal forecasting task} We carefully select a series of typical baselines that have the potential to perform non-consecutive predictions and simultaneously model exogenous context factors~\footnote{As 
the selected methods have overwhelmingly beaten baselines in their corresponding papers, we will not repeatedly compare with the baselines appearing in their papers.}. 
For all baselines, we impose the same retrieval strategy of ours to enable their non-consecutive predictions unless specified, evaluating the effectiveness of  factor-decoupled learning. 
\textbf{(1) Traffic transformer:} An extension of Google's Transformer for traffic forecasting, to capture temporal continuity, periodicity and spatial dependency~\cite{cai2020traffic}.
\textbf{(2) STFGNN:} A framework jointly learns localized heterogeneity and global homogeneity with data-driven graph generation~\cite{li2021spatial}.
\textbf{(3) STG2Seq:} A hierarchical convolutional framework to capture spatial and temporal dependencies of passenger demands with awareness of context factors ~\cite{bai2019stg2seq}.
\textbf{(4) MTGNN:} A state-of-the-art solution to multi-variate time series predictions that can be learned without explicit graph structure~\cite{wu2020connecting}. 
\textbf{(5) ASTGNN:} A state-of-the-art fully-attention based solution to traffic forecasting, by considering contexts of observation themselves~\cite{guo2021learning}. 
\textbf{(6) MTGNN-OSp:} We replace our sampling strategy with the originial  sampling of MTGNN~\cite{wu2020connecting}, to verify the superiority of semantic-neighboring sampling in non-consecutive predictions.

\begin{table*}[]
	\centering
	\caption{Performance comparisons on different settings of grey spatiotemporal systems. We denote $p$-day-a and $q$-day-m as the non-consecutive settings of $p$-day ahead prediction, and $q$-day observation missing in interrupted sensor failures.}
	\label{tab:compar-performance}
	\begin{tabular}{l|llll|llll|llll}
		\toprule
		\hline
		& \multicolumn{4}{c|}{SIP} & \multicolumn{4}{c|}{NYC} & \multicolumn{4}{c}{Metr-LA} \\ \hline
		& 1-day-a & 1-week-a & 3-day-m & 7-day-m & 1-day-a & 1-week-a & 3-day-m & 7-day-m & 1-day-a & 1-week-a & 3-day-m & 7-day-m \\ \hline
		\tabincell{l}{Transformer} & 0.2568 & 0.3012 & 0.2452 & 0.2505 & 0.5542 & 0.5418 & 0.4566 & 0.4752 & 0.2445 & 0.3432 & 0.2275 & 0.3258 \\ \hline
		STFGNN & 0.2588 & 0.3094 & 0.2386 & 0.2634 & 0.2230 & 0.2485 & 0.2310 & 0.2517 & 0.2659 & 0.3655 & 0.2453 & 0.3129 \\ \hline
		STG2Seq & 0.2301 & 0.2873 & 0.2105 & 0.2358 & 0.2027 & 0.2409 & 0.1985 & 0.2035 & 0.3014 & 0.3497 & 0.2894 & 0.3214 \\ \hline
		MTGNN & 0.2458 & 0.2733 & 0.2253 & 0.2406 & 0.2276 & 0.2564 & 0.2134 & 0.2201 & 0.2356 & 0.3025 & 0.2242 & 0.3105 \\ \hline
		ASTGNN & 0.2502 & 0.2788 & 0.2311 & 0.2643 & 0.2345 & 0.2652 & 0.2248 & 0.2351 & 0.2433 & 0.2850 & 0.2301 & 0.2454 \\ \hline
		\tabincell{l}{MTGNN-OSp} & 0.2803 & 0.2917 & 0.2436 & 0.2302 & 0.2517 & 0.2732 & 0.2015 & 0.2115 & 0.2463 & 0.3308 & 0.2461  & 0.2731 \\ \hline
		Ours & \textbf{0.2015} & \textbf{0.2166} & \textbf{0.1726} & \textbf{0.1952} & \textbf{0.1854} & \textbf{0.2023} & \textbf{0.1842} & \textbf{0.1988} & \textbf{0.2022} & \textbf{0.2245} & \textbf{0.2114} & \textbf{0.2405} \\ \hline
		Performance $\uparrow$   & {12.42\%} & {20.76\%} & {18.00\%} & {15.20\%} & {8.50\%} & {16.02\%} & {7.20\%} & {2.31\%} & {14.18\%} & {21.22\%} & {14.10\%} & {2.01\%} \\ \hline
		\bottomrule
	\end{tabular}
\end{table*}

\subsubsection{Uncertainty quantification task}
We reproduce various UQ solutions into two representative forecasting frameworks, STG2Seq and our FDG2S, to illustrate the uncertainty quantification quality. The UQ baselines are as follows, 

\textbf{(1) Dropout BNN:} We realize the Bayesian neural network with dropout~\cite{gal2016dropout}, i.e., plugging dropout layers after graph convolutions in ST learning model.
\textbf{(2) DeepEnsembles:} It trains a series of neural networks with different initializations~\cite{lakshminarayanan2017simple}\footnote{The number of ensembled networks is set as 5, according to~\cite{lakshminarayanan2017simple}. }. 
\textbf{(3) SDE:} It imitates Brownian motions and injects OOD samples~\cite{kong2020sde} into the training process. We modify the loss functions and impose different perturbated samples to realize the SDE.
\textbf{(4) MIS:} It enables a scoring function that rewards narrower confidence or credible intervals and encourages intervals to include the targets.  We implement it with two additional FC layers for upper and lower bound regressions, and combine MAPE with MIS score as an integrated loss~\cite{Wu2021quantify}.
\textbf{(5) STUaNet:} It is an uncertainty-aware spatiotemporal prediction model~\cite{zhou2021stuanet}, and we improve it to multiple steps with LSTM. 

\subsubsection{Metrics}
For \textbf{spatiotemporal learning}, we employ \textbf{MAPE}, as it eliminates the influences of both magnitude orders across datasets and preprocessing strategies across baselines.
For \textbf{uncertainty quantification}, two metrics to jointly quantify the quality of uncertainty quantification. {(1) Prediction interval coverage probability (PICP)}, measures whether the predicted intervals maximally cover the ground-truth, where the higher the value is, the better performance~\cite{wang2019deep}. It can be calculated by, ${\rm PIC{P_{obj}}} = \frac{1}
{{Nh}}\sum\limits_{t = 1}^h {\sum\limits_{i = 1}^N {{b^{i,t}}}}$, 
where $b^{i,t} = 1$ only if the ground-truth is covered by the predicted uncertainty-incorporated intervals $[{\widehat y_{i,t}} - ({\widehat u_{a}})_{i,t}, \; {\widehat y_{i,t}} + ({\widehat u_{a}})_{i,t}]$.
{(2) Uncertainty percentage (UP)}, which measures whether the derived uncertainty is  small enough to avoid infinitely increasing intervals. The lower this metric is, the better~\cite{zhou2021stuanet}, i.e.,
${\rm UP} = \frac{1}
{{Nh}}\sum\limits_{t = 1}^h {\sum\limits_{i = 1}^N {\frac{
({\widehat u_{a}})_{i,t}	
}
		{{{y_{i,t}}}}} }$.
	(3) Performance $\uparrow$: Error decreased ratios compared with best baseline.
Note that only the  aleatoric one is incorporated into interval-level evaluations, as the aleatoric is associated with the potential variations conditioned on factors, while the epistemic one is a relative value  measuring model experiences and it can be  evaluated by illustrating their variations during the training process. 

\subsection{Experimental results}
\subsubsection{Results of spatiotemporal forecasting}
The comparison performances on grey spatiotemporal systems are illustrated in Table~\ref{tab:compar-performance}. Our work consistently and overwhelmingly outperforms baseline methods under two non-consecutive settings. 
We have the following observations. (1) Setting (b) achieves better overall performances than setting (a) and meanwhile performances on 3-day missing  beat those of 7-day missing due to the availability and utilization of nearest observations. These results verify the vital role of nearest observations in spatiotemporal predictions. (2) We delete our semantic-neighboring sampling in MTGNN and the performances heavily deteriorate by at least 6.73\% on 1-week-ahead predictions, which demonstrates the contribution of our semantic-neighboring sampling strategy. (3) For detailed model-specific comparisons, our solution outperforms the best baselines  by 8.50\%$\sim$20.76\% on early planning setting,  and 2.01\%$\sim$18.00\% on  sensor failure setting across three datasets. Traffic transformer and STFGNN achieve barely satisfactory performances on 3-day missing predictions as they are dedicated to traffic forecasting and 
3-day-m is the most similar task to consecutive forecasting. Additionally, STG2Seq and MTGNN perform better on non-consecutive predictions as they are originally designed to involve context vectors. To conclude, competitive results of our FDG2S on non-consecutive forecasting can be attributed to 1) sufficiently exploiting various factors to model initial intensity and aggregation patterns, and 2) the robustness brought by aleatoric uncertainty learning.

\subsubsection{Results of uncertainty quantification}
The results of UQ evaluations are shown in Table~\ref{tab:compar-uq}. 
We observe that our DisEUQ achieves satisfactory quality of interval predictions as it outperforms two state-of-the-art baselines on PICP and obtains comparable UP metric on all datasets. Specifically, dropout-based methods reveal narrow uncertainty intervals but cannot exactly capture the ground-truth in such intervals, while SDE-based solutions achieve a relatively better tradeoff. MIS-based methods reasonably capture the ground-truth with its natural interval-aware objectives but fail to restrict the intervals. In contrast, our DisEUQ can concurrently capture the potential intervals around ground-truths, and prevent the unlimited growth of uncertainty intervals. We also notice that UQ performances vary largely among datasets and methods, which may be because of their unique properties. e.g., For the larger value variations on SIP and fewer fluctuations on Metr-LA, the consistently smaller UPs on Metr-LA than those on SIP are observed. We will further elaborate how  the disentangled uncertainty explains the training process and prediction results in Sec ~\ref{sec:detail_ana}.

\begin{table*}[]
	\small
	\centering
	\caption{Uncertainty quantification comparisons on three datasets}
	\label{tab:compar-uq}
	\begin{tabular}{l|cc|cc|cc}
		\toprule
		\hline
		& \multicolumn{2}{c|}{SIP (PICP $\uparrow$, UP $\downarrow$)}       & \multicolumn{2}{c|}{NYC (PICP $\uparrow$, UP $\downarrow$)}        & \multicolumn{2}{c}{Metr-LA (PICP $\uparrow$, UP $\downarrow$)}    \\ \hline
		& STG2Seq       & Our Net       & STG2Seq       & Our Net        & STG2Seq       & Our Net        \\ \hline
		Dropout BNN   & (0.606,0.368) & (0.618,\textbf{0.419}) & (0.549,0.402) & (0.744,{0.460}) & (0.502,0.335) & (0.754,0.394) \\ \hline
		DeepEnsembles & (0.582,2.580) & (0.697,{1.381}) & (0.524,2.310) & (0.742,2.710)  & (0.628,1.522) & (\textbf{0.876},1.937) \\ \hline
		SDE           & (0.605,\textbf{0.257}) & (0.609,0.587) & (0.615,\textbf{0.233}) & (0.679,0.588)  & (0.677,\textbf{0.235}) & (0.791,0.495) \\ \hline
		MIS           & ({0.652},0.890) & (0.640,1.200)  & (\textbf{0.705},0.965) & (0.714,1.050)  & (0.653,0.645) & (0.720,0.595) \\ \hline
		STUaNet (Multi-step)         & (\textbf{0.657},0.356) & (0.659,0.450)  & ({0.695},0.365) & (0.723,0.468)  & (0.684,0.425) & (0.708,0.423) \\ \hline
		Our UQ        & (0.627,0.386) & (\textbf{0.703},0.494) & (0.507,0.325) & (\textbf{0.754},\textbf{0.450}) & (\textbf{0.688},0.402) & (0.766,\textbf{0.379}) \\ \hline
		\bottomrule
	\end{tabular}
\end{table*}

\subsubsection{Ablation study}
We conduct ablation studies on one-week-ahead predictions, and name ablative variants of FDG2S  as follows.
\textbf{(1)	w/o SF}: remove Status Feature function $\phi_u(y_i, \mathbf{c}(i))$,
\textbf{(2)	w/o TF}: replace the learnable factor-wise  influence decoupling, i.e., Transition Feature function $\phi_p(y_i, y_j|\mathbf{c}(i))$ with a static distance-based adjacent matrix,
\textbf{(3)	w/o STA}: remove the spatiotemporal autoencoder for noise detection,
\textbf{(4)	w/o Var}: remove the spatiotemporal variance of exogenous factor-induced variation.
Table~\ref{tab:ST-ablation} demonstrates the ablation study results. Empirically, the performances of 1-week-ahead predictions deteriorate when two energy functions and Autoencoders are removed. In particular, the factor-wise influence decoupling strategy  plays a most important role to support predictions by the illustrated performance drops of 18\%, 41\% and 17.68\% on three datasets. On the other hand, without ST Variance loss, the performances can be improved, and the reasons may lie in that the introduced uncertainty loss can slightly distract the main objectives.

\subsection{Detailed model analysis and case study}
\label{sec:detail_ana}
\begin{figure}[!]
	\centering
	\includegraphics[scale=0.5]{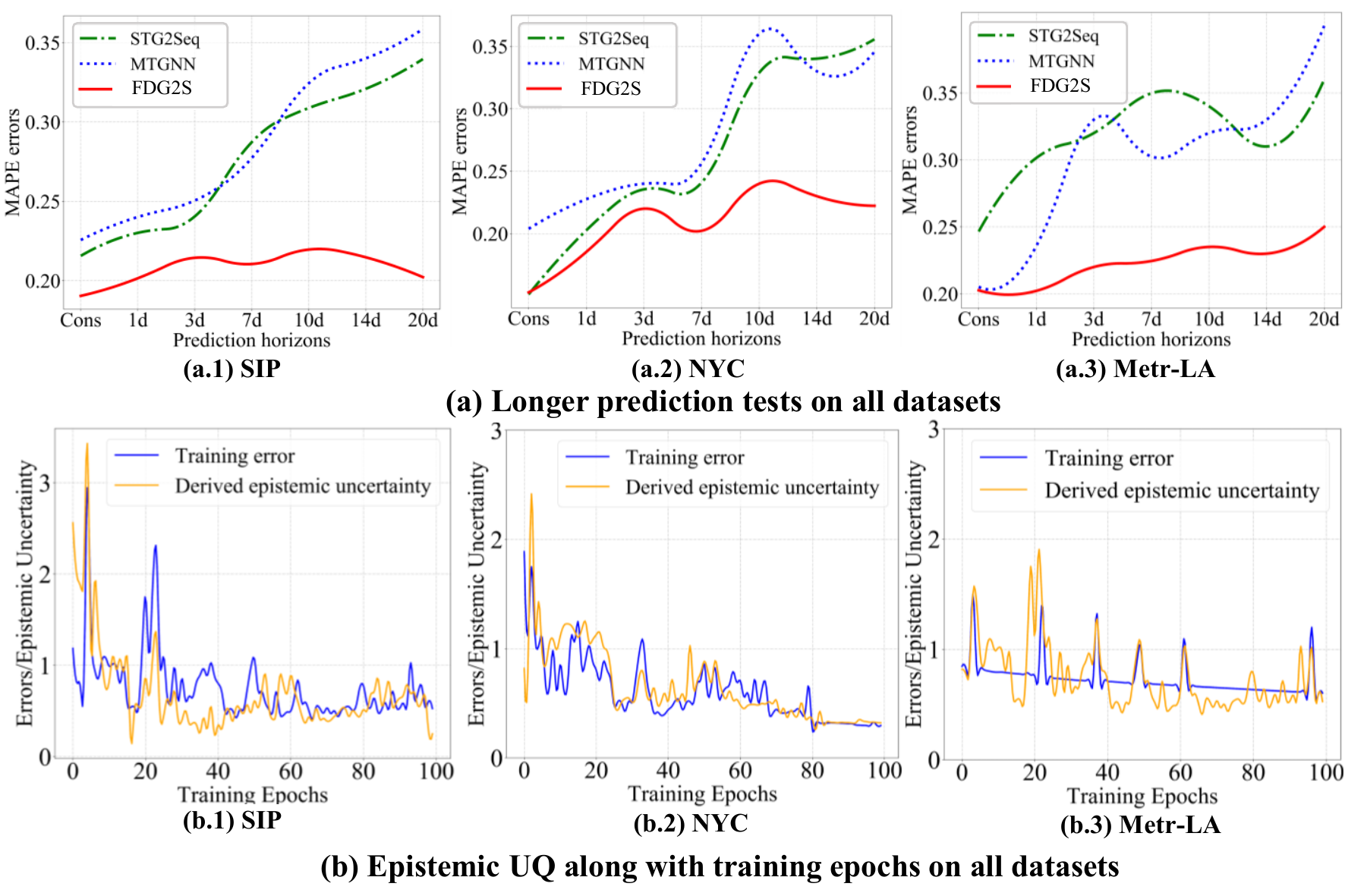}
	\caption{Longer predictions and epistemic UQ results on all datasets. 'Cons' in subfigure(a) refers to performing the next consecutive 6-step predictions.}
	\label{fig:LongPred}
\end{figure}

\textbf{Prediction stability on farther horizons.} We extend the prediction horizons to next 10 days, 14 days, 20 days on three datasets and their performance comparisons with STG2Seq and MTGNN are in Figure~\ref{fig:LongPred}(a). We observe that our FDG2S not only reveals lower errors when predicted horizons become longer, but also performs more stable than others. The superior performances may benefit from three insights, 1) the semantic-based sampling strategy retrieves periodical sequences in a personalized manner, 2) the factor-wise influence decoupling and learnable aggregations for adapting unseen factor combinations, and 3) the reconstruction and weight attenuation constraints of uncertainty quantification. Interestingly, relatively better performances on 7 and 14 days are achieved, probably because of the weekly periodical assistances, which also verifies the intuition of our periodical sampling. 

\textbf{Analysis of epistemic uncertainty. } Here we perform a detailed evaluation on epistemic uncertainty. We present a series of graph-level epistemic uncertainty derived from sample density prober by increasing training epochs in Figure~\ref{fig:LongPred}(b). Epistemic uncertainty results on three datasets vary consistently with errors and tend to decrease with  fluctuations, verifying the rationality of our quantification solution and the inherent regularity of deep learning. These results can promote the interpretability for understanding the prediction reliability of each sample in grey systems, and help re-train or augment critical samples.
In addition, we further demonstrate the consistency between the critical samples and their epistemic uncertainty. Thus,  we filter 5\%, 10\% samples with the highest epistemic uncertainty and then re-train the model. The observed performances on SIP one-week predictions are 19.52\%, 18.99\%, leading promotions of 7\% and 27\%, which confirms that the epistemic uncertainty can exactly indicate the difficulty of model training for each sample. 

\begin{figure}[!]
	\centering
	\includegraphics[scale=0.35]{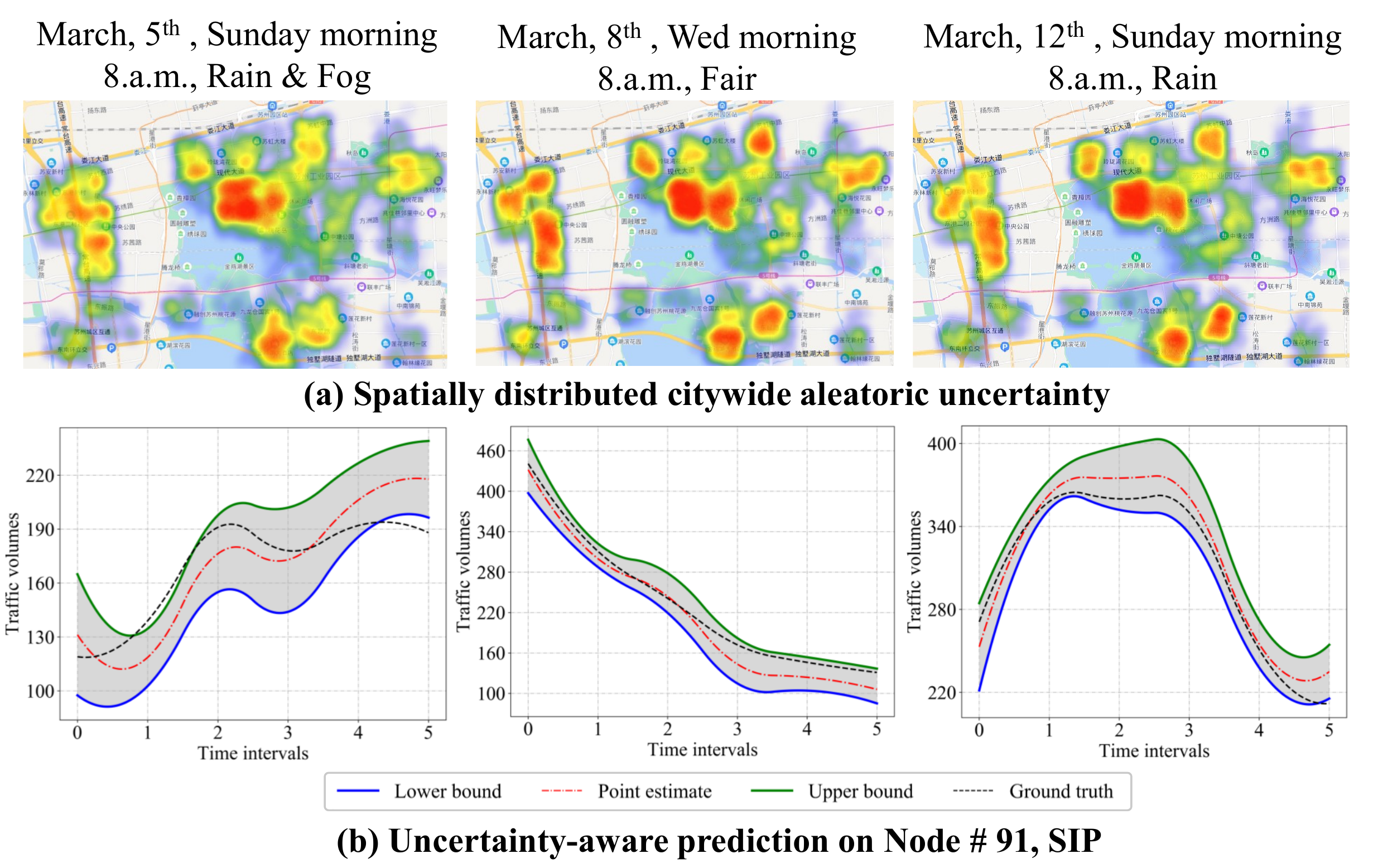}
	\caption{Detailed model analysis and case study}
	\label{fig:Casestuy}
\end{figure}

\textbf{Analysis of aleatoric uncertainty. } 
The spatial maps of aleatoric uncertainty and uncertainty-aware prediction intervals are characterized in Figure~\ref{fig:Casestuy}(a) and (b). 
As illustrated, the aleatoric uncertainty is spatiotemporal heterogeneous, and the uncertainty maps on rainy Sunday are found with similar spatial patterns but both are different from those on fair-weather workdays, verifying the intuition of factor-varying uncertainty. In addition, our predictive intervals tend to be wider when errors become larger, and the widths can exactly capture ground-truth with predicted aleatoric intervals. As a result, these uncertainties can make great sense for  urban perceptions to pre-arrange urban traffic controls and contingency plans during critical activities such as sport competitions and big concerts, avoiding unexpected city emergencies. 

{These promising and rational results jointly demonstrate the robustness and interpretability of our factor-decoupled solution towards learning grey spatiotemporal systems.} 

\subsection{Hyperparameter settings} 
\label{app:hyper}
\begin{figure}[!]
	\centering
	\includegraphics[scale=0.25]{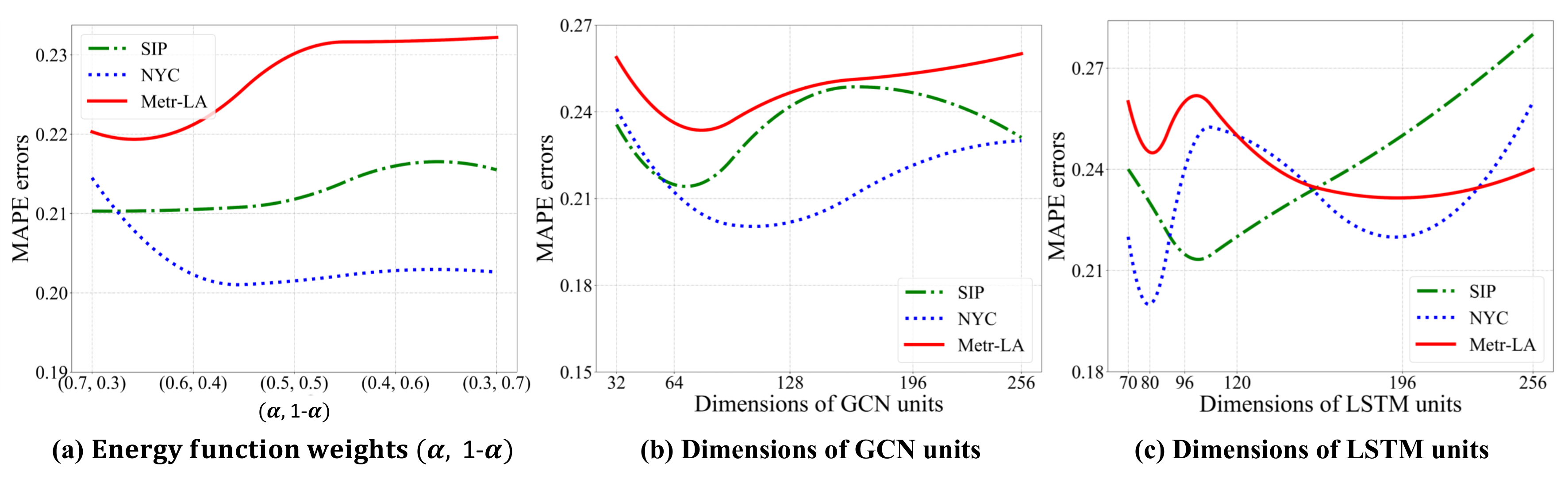}
	\caption{Hyperparameter studies on all datasets}
	\label{fig:Hyper}
\end{figure}

The main hyperparameters are three-fold, the adjustment parameter $\alpha$ concerning two target contributors of CRF energy functions in factor-decoupled aggregation, the dimensions of learnable aggregation kernels and hidden dimensions of LSTM units. We run the hyperparameter searching on one-week predictions to achieve the best performance on each dataset and figure out the process in Figure~\ref{fig:Hyper}.
Since the energy function weights and aggregation kernel dimensions have few influences on performances, we respectively set $\alpha$ to 0.5 and aggregation kernel dimensions to 64 across all datasets. For the hidden dimensions of LSTM, according to Figure~\ref{fig:Hyper}(c),  we set to 96, 80, 196 on SIP, NYC and Metr-LA.

\section{Related Work}
\textbf{Spatiotemporal forecasting.}
Spatiotemporal forecasting receives substantial attention from diverse domains, including traffic~\cite{bai2020adaptive,yu2018spatio,li2017diffusion}, human mobilities~\cite{feng2018deepmove, li2021location}, and smart grids~\cite{wu2020connecting, wu2020adversarial}. Benefiting from non-linear modelling and deep feature extractions, contemporary spatiotemporal forecasting mostly falls into deep learning-based solutions. These methods usually devise various spatial and temporal aggregation or dependence learning strategies, i.e., multi-view GCN~\cite{geng2019spatiotemporal}, spatiotemporal attention~\cite{guo2019attention} and Adaptive GCN-and-GRU ~\cite{bai2020adaptive} to achieve forecasting. 
Among them,  these literature  can be further  classified into immediate fine-grained predictions on hour~\cite{guo2019attention, zhang2017deep} or minute levels~\cite{zhou2020riskoracle,wu2020connecting, bai2020adaptive} and coarse-grained predictions on day levels~\cite{yuan2018hetero, feng2018deepmove}, but both of them make the assumption that they have obtained sufficient  historical observations, especially the recent observations. Unfortunately, in more practical scenarios, early pre-arrangements of both large-scale urban activities and  individual traveling require the model to foresee non-consecutive series, where fragments of observations are unavailable. Promisingly, recent proposals rise the attention of contexts to realize more customized  predictions~\cite{zhang2017deep,bai2019stg2seq,li2021spatial,ye2019co, Bao2019ST}. To this end, the techniques of non-consecutive spatiotemporal forecasting and strategies for decoupling and gathering multi-level factors are highly desired to empower more intelligent urban life.

\textbf{Uncertainty quantification (UQ).}
Uncertainty can be categorized into epistemic and aleatoric types~\cite{der2009aleatory}.  
Regarding epistemic UQ, existing literature aims to capture distributions of model parameters. They devise various techniques, including dropout-based Bayesian neural networks~\cite{liu2020probabilistic,vandal2018quantifying,wang2019deep}, Ensemble methods~\cite{lakshminarayanan2017simple} and Brownian motions~\cite{kong2020sde} to collect multiple outputs from specific inputs, and then leverage sampling-computing strategy to achieve the variance of predictions as epistemic uncertainty. For aleatoric uncertainty, existing works often formulate the uncertainty as the function of inputs and devise  loss functions to  maintain the consistency between errors and learnable aleatoric uncertainty. However, these methods limit uncertainty learning in static images, which are not suitable for spatiotemporal systems.
Pioneering works on numerical weather~\cite{wang2019deep, vandal2018quantifying} and meteorology forecasting~\cite{liu2020probabilistic} initialize the study of spatiotemporal UQ. 
More recently, a state-of-art work~\cite{Wu2021quantify} provides comprehensive baselines and benchmarks on spatiotemporal uncertainty and ~\cite{zhou2021stuanet} devises a variation-based indicator as a guidance for uncertainty learning. Unfortunately, given context-induced heterogeneous uncertainty, these works either neglect considering the model-induced epistemic uncertainty~\cite{zhou2021stuanet}, or fail to internalize the context importance into spatiotemporal uncertainty\cite{vandal2018quantifying,liu2020probabilistic,wang2019deep, Wu2021quantify}, posing challenges to adapting them to our uncertainty quantification of non-consecutive learning.

\textbf{Graph neural networks (GNNs).}
GNNs are firstly proposed  to implement filtering and aggregations of node-level signals along graph topology in spectral domain~\cite{bruna2014spectral}, which suffer the expensive  costs of spectral factorization and the inflexibility of identical neighboring node numbers. Recently, GraphSAGE~\cite{hamilton2017inductive}, GCN~\cite{kipf2016semi}, GAT~\cite{velivckovic2018graph} relax the conditions of neighborhood aggregations and realize series of spatial-based methods.
However, these GNNs focus on fixed graphs with feature-level similarities and cannot adapt to spatiotemporal graphs. To empower an adaptive topology, AGCRN~\cite{bai2020adaptive}  learns a dynamic topology based on feature-wise product, while STAR-GNN~\cite{ma2021improving} and L2P framework ~\cite{xiao2021learning} respectively design a mutual-information based strategy and a generative process, to estimate the optimal receptive fields for GNNs.
More recently, exogenous context factors those are out-of-graph have been demonstrated to potentially interfere the intrinsic graph topology and have complicated interactions on element-wise aggregations~\cite{zhou2020foresee,li2021spatial,guo2021learning}. 
Unfortunately, existing literature has barely discussed this issue.  Even for two seemingly resemble works with ours, i.e., decoupled GNNs~\cite{klicpera2018predict, liu2020towards}  and the disentangled GCN~\cite{ ma2019disentangled}, both of them focus on the disentangling latent fators in graph signals themselves, never considering how to organize or involve the correlated external factors into graph representations. 
\begin{table}[]
	\centering
	\small
	\caption{ Performances on ablative spatiotemporal learning}
	\label{tab:ST-ablation}
	\begin{tabular}{c|c|c|c}
		\toprule
		\hline
		\multirow{2}{*}{Variants}& \multicolumn{3}{c}{MAPE}      \\ \cline{2-4}
		& SIP   & NYC   & Metr-LA \\  \hline
		w/o SF  & 0.2437 & 0.2642 & 0.2856   \\ \hline
		w/o TF  & 0.2499 & 0.2856 & 0.2642   \\  \hline
		w/o STA & 0.2107 & 0.2305 & 0.2305    \\ \hline
		w/o Var & 0.2034 & 0.2128 & 0.2058  \\ \hline
		FDG2S & 0.2105 & 0.2023 & 0.2245  \\ \hline
		\bottomrule
	\end{tabular}
\end{table}

\section{Conclusion}
In this paper, we define a grey spatiotemporal system in which we are desired to perform non-consecutive spatiotemporal forecasting. Technically, we propose a FDG2S by exploiting  exogenous factors for imitating the evolution patterns of unseen future status and providing responsible  uncertainty estimations. Specifically, we first devise a semantic-neighboring sampling to select representative sequences regarding periodicity and instantaneous variations. Then a factor-decoupled aggregation generates basic predictive intensity and performs region-wise aggregations by CRF modelling, enabling adaptive aggregations conditioned on arbitrary factor combinations. Considering the intrinsic incomplete property of grey spatiotemporal systems, we identify two sources of uncertainties  with proposed DisEUQ, which accommodates a sample density prober to explore the epistemic uncertainty regarding regressing capacity, and a weak-supervised aleatoric variation learner to quantify aleatoric uncertainty and attenuate risks from noise and outliers. We evaluate its computational efficiency issue by dedicated analysis and demonstrate the effectiveness of our solution on two non-consecutive settings of G2S. Comprehensive experiments have verified the effectiveness and stability of our FDG2S, and case studies illustrate the interpretability and robustness brought by UQ. For  future works, we will explore a unified model that is adaptive to various non-consecutive settings without model retraining.

\newpage

\bibliographystyle{IEEEtran}
\bibliography{RSTL-kdd}

\end{document}